\documentclass[10pt,a4paper]{IEEEtran}
\usepackage{url}
\usepackage{verbatim}
\usepackage{epsfig}
\usepackage{ifthen}
\usepackage{law}
\usepackage{mystyle}
\usepackage{graphics}
\usepackage{times}
\usepackage{cite}
\usepackage{bbding}
\usepackage{amsfonts}
\usepackage{amsmath}
\usepackage{amssymb}
\usepackage{algorithmic,algorithm}
\usepackage{color}
\usepackage{widetext}
\usepackage{lipsum}

\usepackage[margin=3cm]{geometry}
\usepackage{lipsum,multicol}

\newcommand{\Draft}{0}   

\ifthenelse{\Draft = 1}
{
  \newcommand{\fixMe}[2][]{
    \typeout{***** ERROR: fixMe still in final version *****}
  }
}
{
  \newcommand{\fixMe}[2][] {[{\bf #1}] {\bf \marginpar{\large FIX}} {\em #2}}

}

\newcommand{\bfx}{{\textbf{x}}}

\newcommand{\bfw}{{\textbf{w}}}
\newcommand{\bfy}{{\textbf{y}}}

\title{A Novel Face Recognition Method using Nearest Line Projection
\thanks{Manuscript received August 10, 2011; revised January 2, 2012; accepted April 16, 2012.
\copyright $~$2005 IEEE.}
}

\author{
Huanguo Zhang, Sha Lv,  Wei Li and Xun Qu\\
\normalsize
Department of Electrical Information Engineering, Yibin Vocational \& Technical College\\
Yibin, Sichuan 644003, China}

\date{}
\begin{document}

\maketitle
\thispagestyle{empty}

\begin{abstract}
Face recognition is a popular application of pattern recognition
methods, and it faces challenging problems
including illumination, expression, and pose.
The most popular way is to learn the subspaces of the face images so that it could be project to another
discriminant space where images of different persons can be separated.
In this paper, a nearest line projection algorithm is developed to represent the face images  for face recognition.
Instead of projecting an image to its nearest image, we try to project it to its nearest line spanned by
two different face images.
The subspaces are learned so that each face image to its nearest line is minimized.
We evaluated the proposed algorithm on some benchmark face image database,
and also compared it to some other image projection algorithms.
The experiment results showed that
the proposed algorithm outperforms other ones.
\end{abstract}


\section{Introduction}

In recent years, in the face recognition community,
the manifold based learning methods \cite{jayasekara2013novel,Wang2012,Tang20141,Yao2013563,Wu20131244,Ptucha2013,Vidar2013}
have attracted much attention.
Among these methods, locality preserving projection (LPP) \cite{Li2013100,Huang20132039,Qiao2013280,Hu2007339}
has been the most popular one.
It tries to keep the manifold structure of the image in the
low-dimensional space by mapping
the samples and regularization with a nearest-neighbor graph \cite{Luciska2012254,Crecelius2012952,Liu20131,Chen2013529,Ozaki2013400,Wang20132840}.
Moreover, this method has been improved into handle the non-linear distribution
by the non-linear mapping, and
the locally linear embedding
(LLE) is proposed \cite{Kaiping2013137,Liu2013148,Li2013141}.
However, both these methods are unsupervised \cite{Campello2013344,Li20132559},
which ignore the class label information.
Although they are useful for dimensionality reduction problem \cite{Saeed2013366,Wang2013150},
but they are not suitable for supervised classification problem \cite{Baranowski2013249,Santos20131749}.
To solve this problem, some discriminant subspace learning methods have been proposed, for example,
Linear discriminant analysis  (LDA) \cite{Lu2013165,Zollanvari20133017,Martnezn2013173}, etc.
These methods learn the transformation matrix by
minimizing the intraclass distance  and maximizing interclass distance at the same time \cite{Maliatsos2007,Bala20092198,Ishida2011731}.
Using this criterion, traditional methods such as LPP and LLE can also be extended to
supervised versions.
Moreover, using kernel trick, they can also be extended to kernel versions \cite{Qi20113940,Kung20111807,Ahmed2013562}.

\begin{figure}[!t]
\centering
\includegraphics[width=0.4\textwidth]{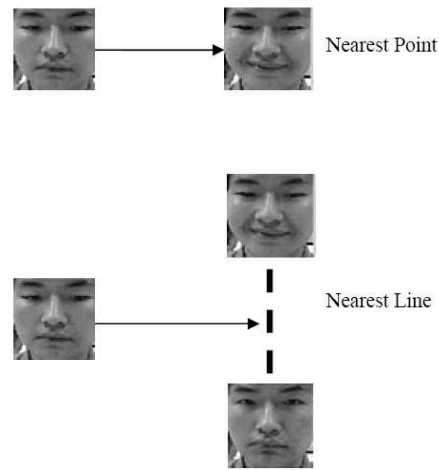}\\
\caption{Nearest Point and Nearest Line.}
\label{Fig1}
\end{figure}

A common feature of all these methods is that they are all using with data points as elements,
and try to keep the locality between
the data points in the new space.
Recently, the nearest linear combination (NLC) method  \cite{Li1998839}
has been proposed.
It treat the line between a pair of data points as the basic elements and
thus
the learning is conducted in
a space spanned by the
data point lines.
To classify a test data point, it is assigned to
its nearest line, instead of its nearest point.
The difference between the nearest point and nearest point is shown in
Figure \ref{Fig1}.
This classification method has been used and encouraging results are given.
But it is only used in classification procedure, and
not used in the feature mapping phase.

In this paper, we propose the
nearest line project for face image mapping problem.
It is different from the traditional data point based
subspace learning which try to map the point into a space where it is close to its nearest
point.
We try to map it so that it is close to its close line.
The transitional methods, including LPP and LLE,
represent the manifold information by constructing the
graph of nearest point.
If the number of data points are small, it could not be a good representation
of the manifold.
To solve this problem, we propose to measure the similarity
between a data point to its nearest lines to explorer
the manifold better.
Moreover, we propose to embed the information between
data point and its nearest line
in the subspace, not in the classification procedure.

This paper is organized as follows. In
Section 2, the proposed algorithm is introduced,
and in Section 3,
experiments are conducted to
compare the
proposed method to other popular projection methods.
In section 4, the paper is concluded.

\section{Nearest Line Projection}

We assume that we have $n$ training data points, denoted as
$\bfx_1,\cdots, \bfx_n$, and
$\bfx_i\in R^d$ is a $d$ dimensional
feature vector for the $i$-th data point.
Subspace project try to map a data point $\bfx$
to a $d'$-dimensional space  by linear projection

\begin{equation}
\begin{aligned}
\bfy = W^\top \bfx
\end{aligned}
\end{equation}
and $W\in R^{d\times d'}$ is the projection matrix.
Usually $d'\ll d$ so that the projection could map the data point into low-dimension space.

\begin{figure}[!t]
\centering
\includegraphics[width=0.5\textwidth]{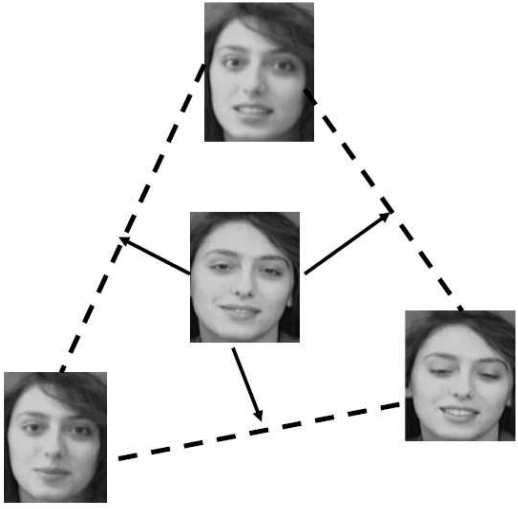}\\
\caption{Mapping  a face image to its nearest  Lines.}
\label{Fig2}
\end{figure}

We hope that after the projection, a data point could be close to its nearest lines. To this end, we need to find it
nearest lines first.
Given a training data point $\bfx_i$, its $K$ nearest neighboring data points $N_i$ are found first

\begin{equation}
\begin{aligned}
N_i=K\arg\min_{j\neq i} \|\bfx_i - \bfx_j\|^2
\end{aligned}
\end{equation}
where $\|\bfx_i - \bfx_j\|^2$ is the $L_2$ norm distance between $\bfx_i$ and $\bfx_j$.
Then we construct the lines from all pairs of data points in $N_i$ as its nearest lines.
In total, there are $C^{N_1}_K$ nearest lines for each data point.
We give an example of finding and mapping a data point to its nearest lines in figure \ref{Fig2}.

\begin{figure*}[!t]
\centering
\includegraphics[width=0.5\textwidth]{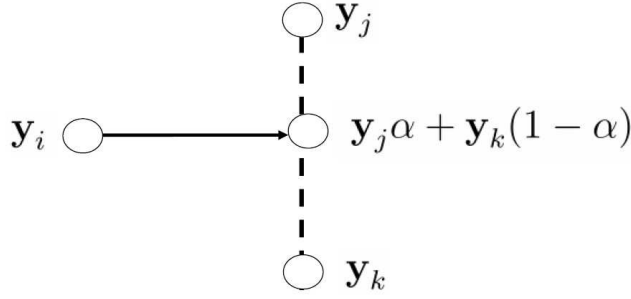}\\
\caption{The distance between $y_i$ and line $L_{jk}(\alpha)$.}
\label{Fig3}
\end{figure*}

Suppose we have
two data points $\bfx_j$ and $\bfx_k$ in $N_i$,
we project them to the low-dimensional space as $\bfy_j$
and $\bfy_k$.
The line between them can be given as

\begin{equation}
\begin{aligned}
L_{jk}(\alpha)=\bfy_j \alpha + \bfy_k (1-\alpha)
\end{aligned}
\end{equation}
and the distance between $\bfy_i$ and this line is computed as
the distance between $\bfy_i$ and its close point at line $L_{jk}(\alpha)$,
as is shown in figure \ref{Fig3}.
To find the close point, we solve the following problem,

\begin{equation}
\min_{\alpha}
\left \{
\begin{aligned}
&D(\alpha)\\
&=\|\bfy_i - L_jk(\alpha)\|^2_2\\
&=
\|\bfy_i - \left[\bfy_j \alpha + \bfy_k (1-\alpha)\right ]\|^2_2\\
&=
\|\bfy_i - \bfy_k -  \alpha (\bfy_j  -\bfy_k)\|^2_2
\end{aligned}
\right \}
\end{equation}
To solve it, we set the derivative of $D(\alpha)$ with regarding to $\alpha$ to zero,

\begin{equation}
\label{equ:alpha}
\begin{aligned}
\frac{\partial D(\alpha)}{\partial \alpha}=
-2 \left [
(\bfy_i - \bfy_k)
-\alpha (\bfy_j  - \bfy_k)
\right ]^\top (\bfy_j - \bfy_k)=0\\
\alpha=
\frac{(\bfy_i - \bfy_k)^\top (\bfy_j - \bfy_k)}{(\bfy_j - \bfy_k)^\top (\bfy_j - \bfy_k)}
\end{aligned}
\end{equation}
By substituting  (\ref{equ:alpha}) to $D(\alpha)$, we have
the distance of a data point $\bfy_i$ to one of its nearest $L_{jk}(\alpha)$ in (\ref{equ:lage1}).

\begin{widetext}
\begin{equation}
\label{equ:lage1}
\begin{aligned}
Dist(\bfy_i, L_{jk};W)
&=\|\bfy_i - \bfy_k -  \alpha (\bfy_j  -\bfy_k)\|^2_2\\
&=
\|\bfy_i - \bfy_k -  \frac{(\bfy_i - \bfy_k)^\top (\bfy_j - \bfy_k)}{(\bfy_j - \bfy_k)^\top (\bfy_j - \bfy_k)}
 (\bfy_j  -\bfy_k)\|^2_2\\
 &=
 \|W^\top\bfx_i - W^\top\bfx_k -  \frac{(W^\top\bfx_i - W^\top\bfx_k)^\top (W^\top\bfx_j - W^\top\bfx_k)}{(W^\top\bfx_j - W^\top\bfx_k)^\top (W^\top\bfx_j - W^\top\bfx_k)}
 (W^\top\bfx_j  -W^\top\bfx_k)\|^2_2\\
&=
 \|W^\top\bfx_i - W^\top\bfx_k -  \frac{(W^\top\bfx_i - W^\top\bfx_k)^\top (W^\top\bfx_j - W^\top\bfx_k)}{(W^\top\bfx_j - W^\top\bfx_k)^\top (W^\top\bfx_j - W^\top\bfx_k)}
 (W^\top\bfx_j  -W^\top\bfx_k)\|^2_2\\
&=
\left
\|W^\top\left ( \bfx_i - \bfx_k -
\frac{(W^\top\bfx_i - W^\top\bfx_k)^\top (W^\top\bfx_j - W^\top\bfx_k)}{(W^\top\bfx_j - W^\top\bfx_k)^\top (W^\top\bfx_j - W^\top\bfx_k)}
 (\bfx_j  -\bfx_k)
 \right )  \right \| ^2_2\\
&=
Tr\left [
W^\top
\left ( \bfx_i - \bfx_k -
\frac{(W^\top\bfx_i - W^\top\bfx_k)^\top (W^\top\bfx_j - W^\top\bfx_k)}{(W^\top\bfx_j - W^\top\bfx_k)^\top (W^\top\bfx_j - W^\top\bfx_k)}
 (\bfx_j  -\bfx_k)
 \right )
 \right. \\
&~~~~~~~~~~~~~~~~~~~\left.
\left ( \bfx_i - \bfx_k -
\frac{(W^\top\bfx_i - W^\top\bfx_k)^\top (W^\top\bfx_j - W^\top\bfx_k)}{(W^\top\bfx_j - W^\top\bfx_k)^\top (W^\top\bfx_j - W^\top\bfx_k)}
 (\bfx_j  -\bfx_k)
 \right )  ^\top
W
\right ]\\
\end{aligned}
\end{equation}
\end{widetext}

We hope to learn a projection matrix $W$ so that the
distance between each data point to its nearest lines could be minimized,
so we obtain the following objective function in (\ref{equ:obje}),

\begin{widetext}
\begin{equation}
\label{equ:obje}
\min_{W}
\left \{
\begin{aligned}
&
\sum_{i=1}^n \sum_{(j,k)\in N_i}
Dist(\bfy_i, L_{jk};W)\\
&=\sum_{i=1}^n \sum_{(j,k)\in N_i}
Tr\left [
W^\top
\left ( \bfx_i - \bfx_k -
\frac{(W^\top\bfx_i - W^\top\bfx_k)^\top (W^\top\bfx_j - W^\top\bfx_k)}{(W^\top\bfx_j - W^\top\bfx_k)^\top (W^\top\bfx_j - W^\top\bfx_k)}
 (\bfx_j  -\bfx_k)
 \right )
 \right. \\
&~~~~~~~~~~~~~~~~~~~\left.
\left ( \bfx_i - \bfx_k -
\frac{(W^\top\bfx_i - W^\top\bfx_k)^\top (W^\top\bfx_j - W^\top\bfx_k)}{(W^\top\bfx_j - W^\top\bfx_k)^\top (W^\top\bfx_j - W^\top\bfx_k)}
 (\bfx_j  -\bfx_k)
 \right )^\top
W
\right ]\\
&=
Tr\left \{
W^\top
\left [
\sum_{i=1}^n \sum_{(j,k)\in N_i}
\left ( \bfx_i - \bfx_k -
\frac{(W^\top\bfx_i - W^\top\bfx_k)^\top (W^\top\bfx_j - W^\top\bfx_k)}{(W^\top\bfx_j - W^\top\bfx_k)^\top (W^\top\bfx_j - W^\top\bfx_k)}
 (\bfx_j  -\bfx_k)
 \right )
 \right. \right. \\
&~~~~~~~~~~~~~~~~~~~\left.\left.
\left ( \bfx_i - \bfx_k -
\frac{(W^\top\bfx_i - W^\top\bfx_k)^\top (W^\top\bfx_j - W^\top\bfx_k)}{(W^\top\bfx_j - W^\top\bfx_k)^\top (W^\top\bfx_j - W^\top\bfx_k)}
 (\bfx_j  -\bfx_k)
 \right )^\top
 \right ]
W
\right \}\\
&=
Tr\left(
W^\top
L(W)
W
\right )
\end{aligned}
\right .
\end{equation}
\end{widetext}
where

\begin{widetext}
\begin{equation}
\begin{aligned}
&L(W)=
\left [
\sum_{i=1}^n \sum_{(j,k)\in N_i}
\left ( \bfx_i - \bfx_k -
\frac{(W^\top\bfx_i - W^\top\bfx_k)^\top (W^\top\bfx_j - W^\top\bfx_k)}{(W^\top\bfx_j - W^\top\bfx_k)^\top (W^\top\bfx_j - W^\top\bfx_k)}
 (\bfx_j  -\bfx_k)
 \right )
 \right. \\
&~~~~~~~~~~~~~~~~~~~\left.
\left ( \bfx_i - \bfx_k -
\frac{(W^\top\bfx_i - W^\top\bfx_k)^\top (W^\top\bfx_j - W^\top\bfx_k)}{(W^\top\bfx_j - W^\top\bfx_k)^\top (W^\top\bfx_j - W^\top\bfx_k)}
 (\bfx_j  -\bfx_k)
 \right )
 \right ]^\top
\end{aligned}
\end{equation}
\end{widetext}

It is obvious that direct optimization of (\ref{equ:obje}) is difficult, so we use the iterative strategy to learn $W$.
In each iteration, we perform the following two steps:

\begin{itemize}
\item
Update $W^{new}$ by fixing $L(W^{old})$:

\begin{equation}
\begin{aligned}
W^{new}=\arg\min_{W}
Tr\left(
W^\top
L(W^{old})
W
\right )
\end{aligned}
\end{equation}
This problem is solved as an eigenvalue problem:

\begin{equation}
\begin{aligned}
L(W^{old})
\bfw
=
\lambda \bfw
\end{aligned}
\end{equation}
where $\bfw$ is the eigenvector of $L$
and $\lambda$ is its corresponding eigenvalue.
We first solve the eigenvectors and eigenvalues,
and then rank the eigenvectors according their eigenvalues
in descending order, and then pick the first $d'$ eigenvectors
to construct the projection matrix $W$:

\begin{equation}
\begin{aligned}
W^{new}=[\bfw_1,\cdots,\bfw_{d'}]
\end{aligned}
\end{equation}

\item
Update $L(W^{new})$ by fixing $W^{new}$:

\begin{equation}
\begin{aligned}
L(W^{new})=L(W)|_{W=W^{new}}
\end{aligned}
\end{equation}
This procedure is directly and simple.
\end{itemize}

Actually this algorithm is within the Expectation maximization (EM) algorithms framework \cite{Covoes20133206,Su2013,Belghith2013876,RodrguezAlvarez20131053}.
Updating $W^{new}$ by fixing $L(W^{old})$ is the maximization step, while
Updating $L(W^{new})$ by fixing $W^{new}$ is the expectation step.

\section{Experiments}

In this section, we perform experiments to compare the proposed algorithms to
other methods,
on the following several face image databases:

\begin{itemize}
\item \textbf{The ORL Database of Faces} \cite{Sakthivel2009367,Deleuze20041110}:
This database is a small database. It only contains 400 images of 40 persons.
For each person, there are 10 images in the database.
The faces in these images are of frontal view and  neutral expression.
Moreover, all the images are captured with  well-controlled conditions, making the recognition quite easy.
However, it has been used very popular in the community of face recognition.
Example images are shown in \ref{Fig4}.

\begin{figure*}[!t]
\centering
\includegraphics[width=0.7\textwidth]{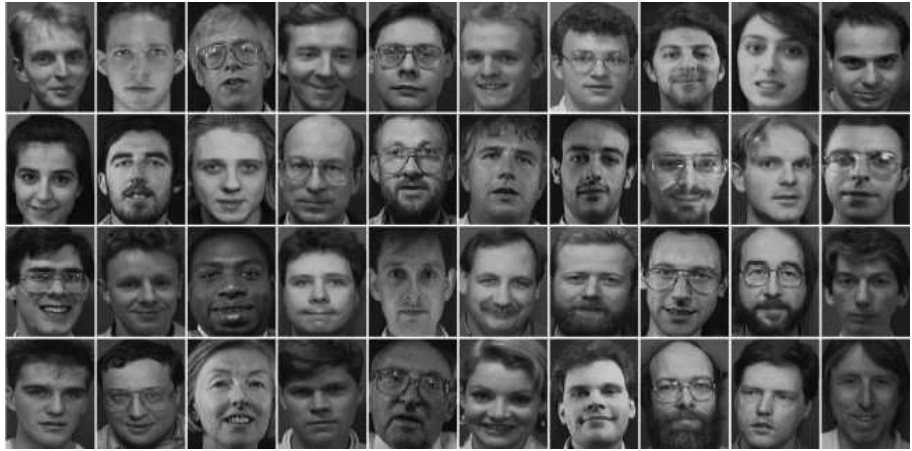}\\
\caption{Face Images in the ORL Database of Faces.}
\label{Fig4}
\end{figure*}

\item \textbf{CMU Database of Faces}
This database is a large scale database, and there are image of 68 persons.
For each person, 170 images are captured.
To make the problem difficult, the images are of different lightning, expression and some faces wear
glasses.
Example images are shown in \ref{Fig5}.

\begin{figure*}[!t]
\centering
\includegraphics[width=0.7\textwidth]{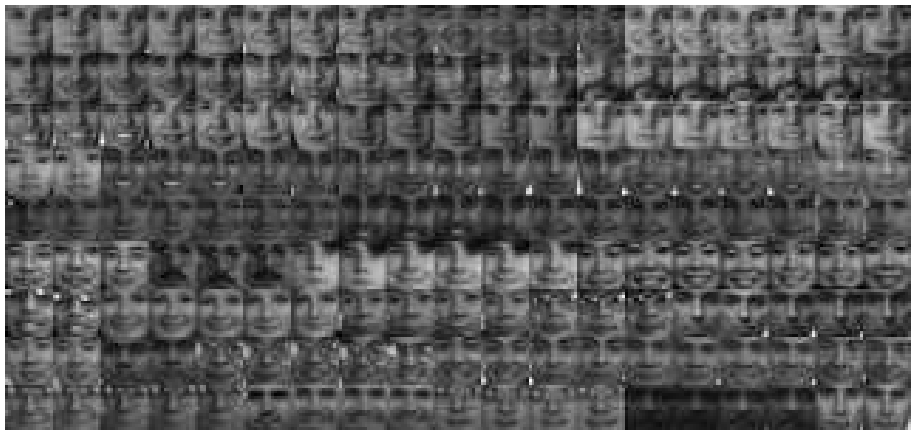}\\
\caption{Face Images in the CMU Database of Faces.}
\label{Fig5}
\end{figure*}

\item \textbf{XM2VTS Database of Faces}
This database is a face image database of 295 persons, and for each person 12 images are collected.
The images are not taken at same time but in different sessions, thus
more variety is introduced.
significant illumination variety are also included.
Example images are shown in \ref{Fig6}.

\begin{figure*}[!t]
\centering
\includegraphics[width=0.5\textwidth]{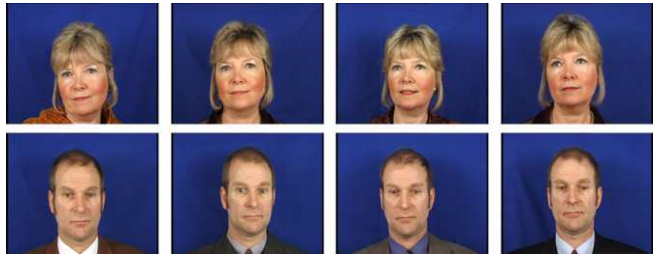}\\
\caption{Face Images in the XM2VTS Database of Faces.}
\label{Fig6}
\end{figure*}

\end{itemize}

\begin{figure}[!t]
\centering
\includegraphics[width=0.5\textwidth]{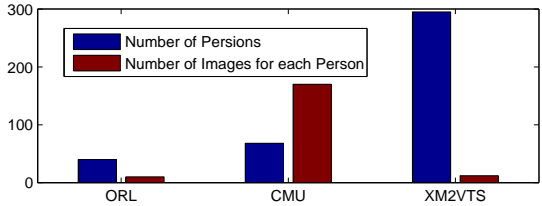}\\
\caption{The statistical information of face image databases.}
\label{Fig7}
\end{figure}

The statistical information of these databases are given in figure \ref{Fig7}

We compare our method against the following methods:

\begin{itemize}
\item Marginal Fisher analysis (MFA) \cite{Sun20125897,Liu20131025,Liu20132051,Jiang2013};

\item
Locality preserving projection (LPP) \cite{Li2013100,Huang20132039,Qiao2013280,Hu2007339};

\item
Neighborhood preserving discriminant analysis (NPDA) \cite{Lu20103447,Lin20111254,Liang2011571,Wang20112464}.

\end{itemize}
We denote the proposed nearest line projection as (LNP).

In the experiment, we split the database into two subsets --- training set and test set randomly,
and the split are performed for 10 times, and the average recognition rates are reported as the results.
For classification, we used the nearest neighbor classifier \cite{Lakhotia2013109,Dhurandhar2013259,Khoa2013,Wang2013136}.

\begin{figure}[!t]
\centering
\includegraphics[width=0.5\textwidth]{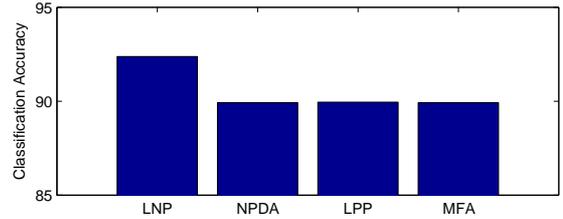}\\
\caption{Experiment results on ORL database.}
\label{Fig8}
\end{figure}

\begin{figure}[!t]
\centering
\includegraphics[width=0.5\textwidth]{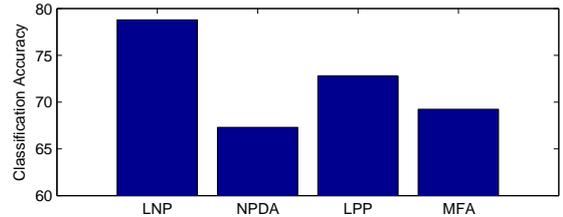}\\
\caption{Experiment results on CMU database.}
\label{Fig9}
\end{figure}

\begin{figure}[!t]
\centering
\includegraphics[width=0.5\textwidth]{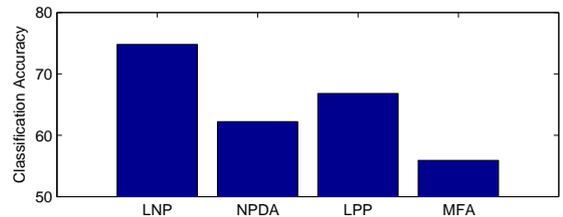}\\
\caption{Experiment results on XM2VTS database.}
\label{Fig10}
\end{figure}

The experiment results on different databases are reported in
figure \ref{Fig8}, \ref{Fig9},  and \ref{Fig10}.
From these figures, we can see that the performance of the NLP is always better than that of the other
algorithms.
This is an strong evidence that the proposed nearest line based algorithm can outperform the
data point based projection methods.

\section{Conclusion}

In this paper, we proposed a novel data projection
by mapping the data points to its nearest lines.
Instead of using data points as
projection elements, we used the
line spanned by two different points to project the data
points.
An iterative algorithm is developed for the projection problem.
The encouraging results showed the advantage of the proposed method.
In the future, we will investigate the probability of using the line projection
to the matrix factorization problems.

\end{document}